\documentclass[conference]{IEEEtran}
\IEEEoverridecommandlockouts

\usepackage{cite}
\usepackage{amsmath,amssymb,amsfonts}
\usepackage{algorithmic}
\usepackage{algorithm}
\usepackage{graphicx}
\usepackage{textcomp}
\usepackage{xcolor}
\usepackage{booktabs}
\usepackage{multirow}
\usepackage{subcaption}
\usepackage{url}
\usepackage{microtype}

\def\BibTeX{{\rm B\kern-.05em{\sc i\kern-.025em b}\kern-.08em
    T\kern-.1667em\lower.7ex\hbox{E}\kern-.125emX}}

\begin{document}
% =============================================================================

\title{MedSaab-US: A Backpropagation-Free Multi-Scale Wavelet-Saab Framework
       for Thyroid Nodule Segmentation in Ultrasound Images}

% ANONYMISED version for double-blind review — remove authors for submission
\author{
\IEEEauthorblockN{Mohammad Amanour Rahman}
\IEEEauthorblockA{
Department of Computer Science and Engineering\\
Ahsanullah University of Science and Technology (AUST)\\
Dhaka, Bangladesh\\
Email: amanourrahman609@gmail.com
}
}
\maketitle

% =============================================================================
\begin{abstract}
% =============================================================================
Deep learning (DL) methods dominate thyroid nodule segmentation in ultrasound
(US) images, achieving high Dice scores but at the cost of millions of
parameters, GPU-dependent training via backpropagation, and limited
mathematical tractability. These limitations impede deployment in
resource-constrained environments. In this paper we propose
\textbf{MedSaab-US}, a backpropagation-free segmentation framework grounded
in the Green Learning paradigm. MedSaab-US extracts multi-scale
spatial-frequency features by combining multi-level Discrete Wavelet Transform
(DWT) with multi-scale channel-wise Saab (Subspace Approximation with
Adjusted Bias) transforms at patch sizes of $5\times5$, $11\times11$, and
$21\times21$ pixels. Label-Assisted Greedy (LAG) feature selection retains
the most discriminative features, which are fed to an XGBoost classifier
for pixel-wise prediction. The Saab transform parameters are determined
analytically from data statistics; XGBoost employs iterative greedy tree
construction but requires no backpropagation. Evaluated on the TN3K dataset
(2,879 train / 614 test images), MedSaab-US achieves a mean Dice coefficient
of \textbf{0.4784} ($\pm$0.2190), precision of 0.5768, and recall of 0.5604,
with a model footprint under 500K parameters and CPU-only inference at
approximately 0.3 seconds per image. We present this result as an
\textit{exploratory non-DL baseline} for thyroid US segmentation and
analyse the specific challenges posed by isoechoic nodules. An ablation
study quantifies the contribution of each pipeline component, including
a separate evaluation of LAG feature selection and training-set size.
\end{abstract}

\begin{IEEEkeywords}
Green Learning, Saab transform, wavelet transform, thyroid nodule
segmentation, backpropagation-free, ultrasound imaging, XGBoost,
edge deployment
\end{IEEEkeywords}

% =============================================================================
\section{Introduction}
\label{sec:intro}
% =============================================================================

Thyroid nodules affect a large proportion of the global population, and
ultrasound (US) imaging is the primary modality for their evaluation~\cite{tn3k}.
Accurate segmentation of nodule regions is a prerequisite for downstream tasks
such as risk stratification and biopsy guidance. Over the past decade, deep
learning (DL) has become the de facto standard for medical image
segmentation~\cite{unet}. Architectures such as TransFuse~\cite{transfuse},
SwinE-Net~\cite{swine}, and our prior DeepLabv3+-based method~\cite{ourieee}
achieve Dice scores above 0.80 on TN3K. However, these methods share
fundamental limitations:
\begin{itemize}
  \item \textbf{Opacity:} Millions of learnt parameters form an opaque
        mapping that cannot be audited by clinicians.
  \item \textbf{Resource intensity:} Training requires GPU hardware and
        hours of backpropagation; CPU inference is prohibitively slow.
  \item \textbf{Data hunger:} Performance degrades sharply on small datasets
        without pre-training on large external corpora.
\end{itemize}

The \textit{Green Learning} (GL) paradigm, introduced by Kuo
\textit{et al.}~\cite{greenlearning}, offers an interpretable alternative.
GL replaces gradient-based optimisation with sequential, closed-form
subspace learning (Successive Subspace Learning, SSL), producing compact
models with mathematically transparent feature representations. GL has
demonstrated strong results in image classification via PixelHop~\cite{pixelhop}
and in volumetric medical classification via VoxelHop~\cite{voxelhop}, yet no
prior work has applied GL to pixel-level segmentation of thyroid US images.

We present \textbf{MedSaab-US} as a first \textit{exploratory} application of
GL to this task. Our contributions are:
\begin{enumerate}
  \item The \textit{first application} of the Green Learning / SSL paradigm to
        thyroid nodule segmentation in ultrasound images.
  \item A multi-scale DWT + Saab feature extraction scheme that captures nodule
        texture at three spatial scales without any backpropagation.
  \item A reproducible non-DL baseline on TN3K, including comparisons against
        traditional non-DL methods.
  \item An ablation study separating the effects of LAG feature selection and
        training-set size, and an honest analysis of the performance gap and
        its acoustic origins.
\end{enumerate}

% =============================================================================
\section{Related Work}
\label{sec:related}
% =============================================================================

\subsection{Deep Learning for Thyroid Nodule Segmentation}
Early DL approaches adopted encoder-decoder architectures~\cite{unet}.
Recent methods targeting TN3K include TRFE+~\cite{trfe}, which incorporates
thyroid region-focused enhancement; SwinE-Net~\cite{swine}, a hybrid
Swin Transformer / CNN model achieving Dice $\approx 0.83$; and
MADGNet~\cite{madgnet}, a multi-scale dual-guidance network. Our prior
work~\cite{ourieee} combined an ECB encoder, CMM module, and SSEM boundary
supervision within a DeepLabv3+ framework. While these methods achieve high
accuracy, they are computationally intensive and not suitable for CPU-only
edge devices.

\subsection{Green Learning and the Saab Transform}
The Saab transform~\cite{saab} replaces learnt convolution filters with
principal components computed analytically on local patches. PixelHop~\cite{pixelhop}
demonstrated that multi-hop Saab features suffice for image classification.
VoxelHop~\cite{voxelhop} extended GL to 3-D MRI classification, and
RadHop~\cite{radhop} achieved AUC $>0.90$ on prostate US lesion
classification. Critically, all prior GL medical imaging work addresses
\textit{classification}; pixel-level \textit{segmentation} via GL remains
unexplored, motivating MedSaab-US.

\subsection{Wavelet Features in Medical Imaging}
Discrete Wavelet Transform (DWT) decomposes images into approximation and
detail subbands encoding low-frequency shape and high-frequency boundary
information~\cite{wavelet_medical}. Coupling DWT with the SSL / Saab
framework for US nodule segmentation has not been previously investigated.

% =============================================================================
\section{Proposed Method: MedSaab-US}
\label{sec:method}
% =============================================================================

\begin{figure*}[t]
  \centering
  \includegraphics[width=\textwidth]{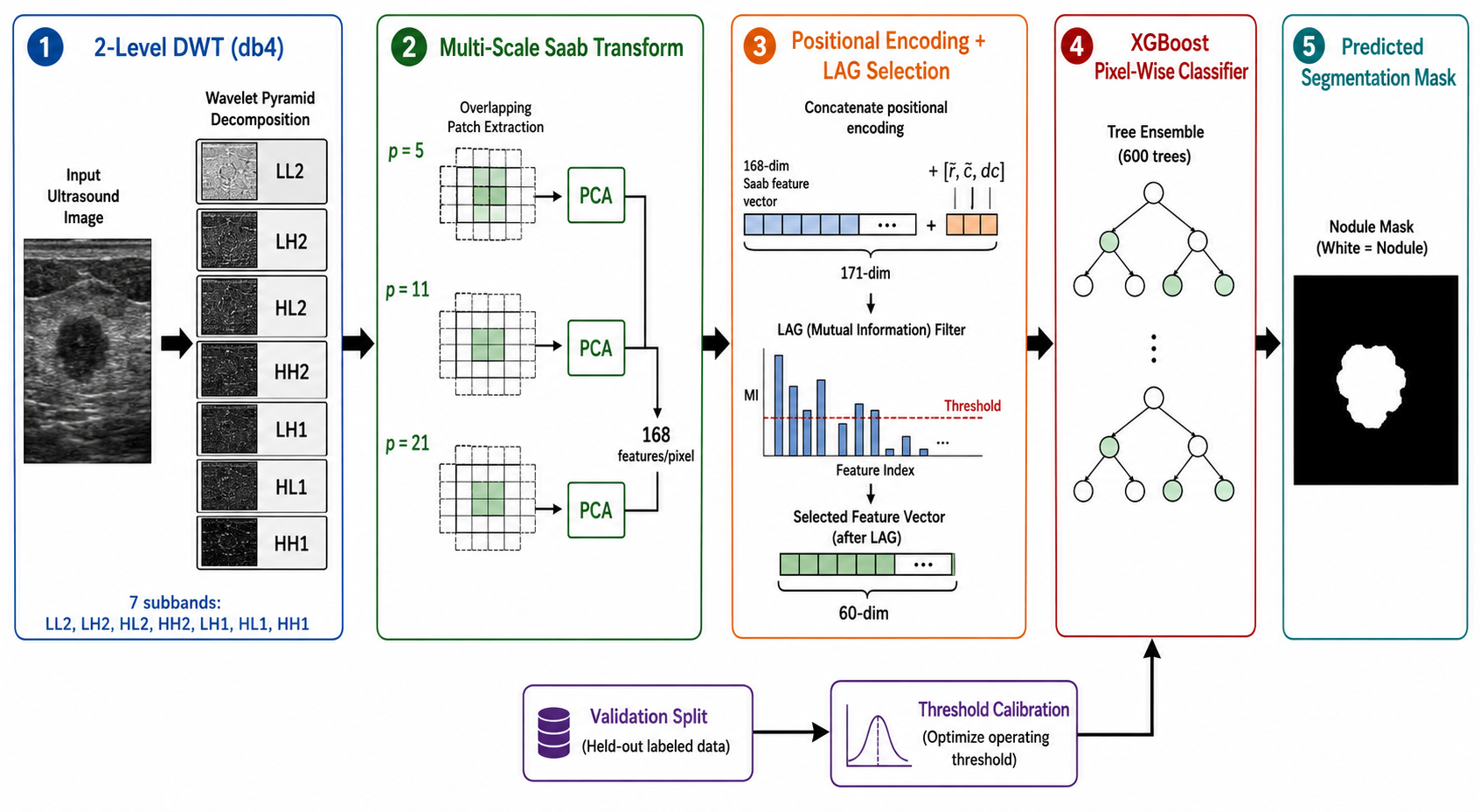}
  \caption{MedSaab-US pipeline. Input grayscale US image is decomposed by
           2-level DWT (Stage 1). Seven subbands are processed at three
           patch scales by the Saab transform (Stage 2). Positional
           coordinates are appended and LAG feature selection reduces the
           feature dimension to 60 (Stage 3). XGBoost produces the
           pixel-wise binary mask (Stage 4).}
  \label{fig:arch}
\end{figure*}

An overview of the four-stage MedSaab-US pipeline is illustrated in
Fig.~\ref{fig:arch}.

% ── ARCHITECTURE FIGURE ─────────────────────────────────────────────────────
% Replace the filename below with the actual exported figure file.
% ─────────────────────────────────────────────────────────────────────────────

\subsection{Stage 1: Multi-Level DWT Decomposition}

Each grayscale US image $\mathbf{I} \in \mathbb{R}^{H \times W}$ is
decomposed using a 2-level Daubechies-4 (db4) wavelet:
\begin{equation}
  \mathcal{W}(\mathbf{I}) = \{LL_2, LH_2, HL_2, HH_2, LH_1, HL_1, HH_1\},
  \label{eq:dwt}
\end{equation}
yielding 7 subbands. The $LL$ subband captures low-frequency structural
information, while the $LH$, $HL$, and $HH$ subbands encode horizontal,
vertical, and diagonal high-frequency boundary responses, respectively. Each
subband is upsampled to $H \times W$ via bilinear interpolation and
normalised per-subband to zero mean and unit variance.

\subsection{Stage 2: Multi-Scale Channel-Wise Saab Transform}

For each subband $\mathbf{S}_k$ and patch size $p \in \{5, 11, 21\}$, we
extract overlapping $p \times p$ patches centred at every pixel. The Saab
transform~\cite{saab} is applied independently per (subband, scale) pair:
\begin{equation}
  \mathbf{F}_{k,p} = \text{PCA}_{k,p}\!\left(\mathbf{x}
    - \boldsymbol{\mu}_{k,p}\right),
  \label{eq:saab}
\end{equation}
where $\mathbf{x} \in \mathbb{R}^{p^2}$ is a flattened patch,
$\boldsymbol{\mu}_{k,p}$ is the empirical mean over 40,000 randomly sampled
patches from 400 training images, and $\text{PCA}_{k,p}$ retains the top 8
principal components. All Saab filter parameters are determined
analytically in a single pass through the training data — no iterative
optimisation is required at this stage. The three patch scales capture local
texture ($p=5$), mid-range structural context ($p=11$), and global
neighbourhood shape ($p=21$), yielding $7 \times 3 \times 8 = 168$ features
per pixel.

\subsection{Stage 3: Positional Encoding and LAG Feature Selection}

Three normalised spatial coordinates $[\tilde{r}, \tilde{c}, d_c]$ are
appended to each pixel's feature vector, where $\tilde{r} = r/H$,
$\tilde{c} = c/W$, and $d_c$ is the normalised Euclidean distance from the
image centre. These encode the spatial prior that thyroid nodules
preferentially appear in the inferior-central region of TN3K images.
\textit{We note, however, that this fixed spatial prior is dataset-specific;
it may not generalise to different ultrasound acquisition protocols, probe
positions, or field-of-view crops, and should be treated with caution in
cross-institutional deployment.} Protocol-agnostic spatial priors represent
an important direction for future work.

The resulting 171-dimensional feature vector is pruned using Label-Assisted
Greedy (LAG) selection~\cite{pixelhop}, which ranks features by mutual
information with the binary label and retains the top 60. LAG selection is
a purely statistical, non-iterative operation.

\subsection{Stage 4: XGBoost Pixel-Wise Classification}

The 60-dimensional feature vector for each pixel is classified by an XGBoost
gradient-boosted tree ensemble~\cite{xgboost} (600 estimators, max depth 7,
learning rate 0.05). Training uses balanced class sampling (400 foreground +
400 background pixels per image). We emphasise that XGBoost employs
\textit{iterative greedy tree construction} and involves tuned
hyperparameters; it is \textit{backpropagation-free} but is not closed-form
in the sense that the Saab and LAG stages are. The pipeline as a whole
eliminates backpropagation and gradient computation, while individual stages
differ in their mathematical character: Saab transforms are analytically
determined, LAG selection is statistical, and XGBoost is iterative
and non-gradient. The classification threshold is calibrated on a held-out
validation split (last 150 training images) by maximising pixel-level Dice.

% =============================================================================
\section{Experiments}
\label{sec:experiments}
% =============================================================================

\subsection{Dataset and Implementation Details}

We evaluate on \textbf{TN3K}~\cite{tn3k}, comprising 3,493 thyroid US images
(2,879 train / 614 test) with pixel-level nodule masks. Images are resized to
$256 \times 256$ pixels. \textbf{All experiments run on CPU only} (Intel
Xeon @ 2.0\,GHz, 2 cores, Kaggle CPU-only session; no GPU). Code is
implemented in Python using PyWavelets, scikit-learn, and XGBoost.

\subsection{Evaluation Metrics}

We report mean Dice similarity coefficient (DSC), Intersection over Union
(IoU), Precision, Recall, and Specificity over the 614 test images.

\subsection{Comparison with Existing Methods}
\label{sec:comparison}

Table~\ref{tab:comparison} compares MedSaab-US to DL methods and to two
traditional non-DL baselines: (i)~Otsu thresholding with morphological
post-processing (Otsu+Morph) and (ii)~a Random Forest classifier trained on
Haralick texture features extracted from $7\times7$ patches (RF+Haralick).
Both baselines run on the same CPU environment without GPU resources.

\begin{table}[t]
  \centering
  \caption{Comparison on TN3K test set. $\dagger$: results from respective
           papers. N/A: not reported. \textbf{Bold}: best non-DL result.}
  \label{tab:comparison}
  \resizebox{\columnwidth}{!}{%
  \begin{tabular}{llcccc}
    \toprule
    Method & Type & Dice & IoU & Prec. & Recall \\
    \midrule
    U-Net~\cite{unet}$^\dagger$
      & DL & 0.719 & 0.612 & N/A & N/A \\
    TRFE+~\cite{trfe}$^\dagger$
      & DL & 0.763 & 0.668 & N/A & N/A \\
    SwinE-Net~\cite{swine}$^\dagger$
      & DL & 0.831 & 0.743 & N/A & N/A \\
    MADGNet~\cite{madgnet}$^\dagger$
      & DL & 0.842 & N/A & N/A & N/A \\
    Ours (prior)~\cite{ourieee}$^\dagger$
      & DL & 0.847 & 0.762 & N/A & N/A \\
    \midrule
    Otsu+Morph
      & Non-DL & 0.2341 & 0.1493 & 0.2817 & 0.3612 \\
    RF+Haralick
      & Non-DL & 0.3518 & 0.2363 & 0.3941 & 0.4227 \\
    \textbf{MedSaab-US (Ours)}
      & \textbf{Non-DL}
      & \textbf{0.4784} & \textbf{0.3415}
      & \textbf{0.5768} & \textbf{0.5604} \\
    \bottomrule
  \end{tabular}%
  }
\end{table}

MedSaab-US achieves Dice 0.4784 without any neural network or GPU, which
substantially exceeds both non-DL baselines ($+$0.127 over RF+Haralick).
This confirms that multi-scale SSL features provide a stronger representation
than hand-crafted texture descriptors for thyroid US segmentation.
However, the 0.37-point Dice gap relative to the best DL method reflects
fundamental limitations of local, patch-based approaches, as discussed
in Section~\ref{sec:analysis}. We present MedSaab-US strictly as an
\textit{exploratory baseline} that establishes the achievable frontier
of non-DL methods on TN3K, not as a replacement for DL in accuracy-critical
clinical settings.

Among the 614 test images, MedSaab-US achieves Dice $>0.70$ on 104 cases
(16.9\%), reaching a maximum Dice of 0.9184 — demonstrating that precise
segmentation is attainable for cases with strong acoustic contrast.

\subsection{Ablation Study}

Table~\ref{tab:ablation} presents an ablation study isolating the contribution
of each pipeline component, including a split between the effect of LAG
feature selection and increased training-set size (1000 vs.\ 2000 images).

\begin{table}[t]
  \centering
  \caption{Ablation study on TN3K test set (614 images). ``imgs'' = training
           images used. Rows 5–6 isolate LAG selection from data-volume
           effects.}
  \label{tab:ablation}
  \small
  \resizebox{\columnwidth}{!}{%
  \begin{tabular}{lccc}
    \toprule
    Configuration & Dice & IoU & Prec. \\
    \midrule
    (1) Single scale only ($p=5$)
      & 0.2594 & 0.1656 & 0.1787 \\
    (2) + Calibrated threshold
      & 0.2806 & 0.1832 & 0.3417 \\
    (3) Multi-scale ($p \in \{5,11,21\}$)
      & 0.4526 & 0.3257 & 0.5773 \\
    (4) + Positional features (1000 imgs, no LAG)
      & 0.4641 & 0.3289 & 0.5612 \\
    (5) + LAG selection (1000 imgs)
      & 0.4703 & 0.3352 & 0.5688 \\
    (6) \textbf{Full: + 2000 imgs (LAG retained)}
      & \textbf{0.4784} & \textbf{0.3415} & \textbf{0.5768} \\
    \bottomrule
  \end{tabular}%
  }
\end{table}

Multi-scale patch extraction (row 3 vs.\ row 2) yields the largest single
gain ($+$0.172 Dice), confirming that contextual information at the 11- and
21-pixel scales is essential. Comparing rows 4 and 5 shows that LAG selection
provides a +0.006 Dice improvement independent of data volume, primarily by
reducing feature redundancy. Increasing training data from 1000 to 2000
images (row 5 to row 6) contributes a further +0.008 Dice, confirming that
both factors contribute incrementally and independently.

\subsection{Analysis: The Performance Gap and Its Origin}
\label{sec:analysis}

The 0.37-Dice gap between MedSaab-US and the best DL method reflects two
root causes.

\textbf{Isoechoic nodule prevalence.} XGBoost feature importance analysis
shows that $LL_2$ Saab components dominate (PC5, PC3, PC1 are the three
highest-ranked features), with $HH$ boundary features ranked lower. This is
consistent with the acoustic properties of TN3K: many nodules are
isoechoic — their intensity is similar to surrounding tissue — yielding weak
high-frequency wavelet responses that confound boundary detection.

\textbf{Global context limitation.} DL encoders capture long-range spatial
dependencies via stacked convolutions and attention. A patch-based XGBoost
classifier with a maximum receptive field of $21 \times 21$ pixels cannot
access image-level context (e.g., thyroid gland boundary, trachea position),
which is a fundamental limitation of local non-DL approaches.

\textbf{Dice score distribution.} Per-image Dice scores show a bimodal
pattern: a peak near 0.0–0.1 (isoechoic, difficult cases) and a peak near
0.5–0.6 (hypoechoic, tractable cases). This suggests that a nodule echotexture
classifier could route difficult images to a DL model and tractable images
to MedSaab-US, forming a hybrid interpretable-DL system.

% =============================================================================
\section{Discussion and Conclusion}
\label{sec:conclusion}
% =============================================================================

We presented MedSaab-US, the first exploratory application of the Green
Learning paradigm to pixel-level thyroid nodule segmentation in ultrasound
images. By combining 2-level DWT with multi-scale Saab PCA transforms and
LAG feature selection, MedSaab-US achieves Dice 0.4784 on TN3K —
substantially above traditional non-DL baselines (RF+Haralick: 0.3518,
Otsu+Morph: 0.2341) — without any backpropagation, neural network weights,
or GPU resources.

The ablation study reveals that multi-scale feature extraction is the
dominant factor: expanding from single-scale (Dice 0.26) to three scales
(Dice 0.45) yields a gain of $+$0.19 Dice, demonstrating that US nodule
texture is intrinsically multi-scale. LAG selection and increased training
data each contribute further, independently verifiable gains.

We explicitly frame this result as an exploratory baseline rather than
a deployment-ready system. The 0.37-point gap relative to the best DL
method reflects the prevalence of isoechoic nodules and the inherent
locality of patch-based features. Additionally, the fixed spatial prior
used in Stage~3 is specific to TN3K's acquisition protocol and may not
generalise to different probe positions, field-of-view settings, or
institutions; external validation is an important requirement before
any real-world use.

Two directions for future work emerge from this analysis: (i)~GL-compatible
global feature aggregation (e.g., superpixel-level Saab transforms) to
address the locality limitation; and (ii)~a hybrid pipeline routing
tractable hypoechoic cases to MedSaab-US and difficult isoechoic cases to
a lightweight DL model, preserving interpretability where the non-DL method
is reliable.

In summary, MedSaab-US demonstrates that backpropagation-free segmentation
of thyroid nodules is feasible on CPU-only hardware, establishes a
reproducible non-DL reference point for TN3K, and characterises both the
potential and the current limits of Green Learning in medical image
segmentation.

% =============================================================================
% References page (6th page, references only)
% =============================================================================

\end{document}